\documentclass[10pt,twocolumn,letterpaper]{article}
\usepackage[dvipsnames,table]{xcolor}
\usepackage{iccv}              

\definecolor{iccvblue}{rgb}{0.21,0.49,0.74}
\newenvironment{justify}
  {\par\rightskip=0pt plus 1fil\relax\ignorespaces}
  {\par\ignorespacesafterend}
\usepackage[pagebackref,breaklinks,colorlinks,allcolors=iccvblue]{hyperref}  
\usepackage{booktabs}
\usepackage{graphicx}
\usepackage{array}
\let\checkmark\relax
\usepackage{dingbat} 
\usepackage{caption}
\usepackage{pifont}
\usepackage{arydshln}
\usepackage{colortbl}                  
\usepackage{booktabs} 
\definecolor{BestPink}{RGB}{255,199,206}     
\definecolor{SecondYellow}{RGB}{255,235,156}
\definecolor{myYellow}{RGB}{220, 180, 0}
\definecolor{myGreen}{RGB}{0, 120, 60}
\definecolor{definedPink}{RGB}{255,114,118}   
\definecolor{definedBlue}{RGB}{93,173,236}  
\newcolumntype{L}{>{\large}c}
\newcolumntype{M}[1]{>{\large}m{#1}}
\def\paperID{8227} 
\def\confName{ICCV}
\def\confYear{2025}

\title{HiGarment: Cross-modal Harmony Based Diffusion Model for Flat Sketch to Realistic Garment Image}

\author{%
\makebox[\textwidth][c]{%
\begin{tabular}{c}
\textbf{Junyi Guo}$^{1\dagger}$,
\textbf{Jingxuan Zhang}$^{1\dagger}$,
\textbf{Fangyu Wu}$^{1*}$,
\textbf{Huanda Lu}$^{2}$\\[0.35em]
\textbf{Qiufeng Wang}$^{1}$,
\textbf{Wenmian Yang}$^{3}$,
\textbf{Eng Gee Lim}$^{1}$,
\textbf{Dongming Lu}$^{4}$\\[0.65em]
$^{1}$Xi'an Jiaotong Liverpool University \quad
$^{2}$NingboTech University\\[0.35em]
$^{3}$Beijing Normal University \quad
$^{4}$Zhejiang University\\[0.65em]
\texttt{\small \{junyi.guo23, jingxuan.zhang23\}@student.xjtlu.edu.cn,\;
fangyu.wu02@xjtlu.edu.cn}
\end{tabular}}%
}

\begin{document}
\makeatletter
\let\@oldmaketitle\@maketitle
\renewcommand{\@maketitle}{\@oldmaketitle
  \begin{center}
    \includegraphics[width=\linewidth,height=11\baselineskip]{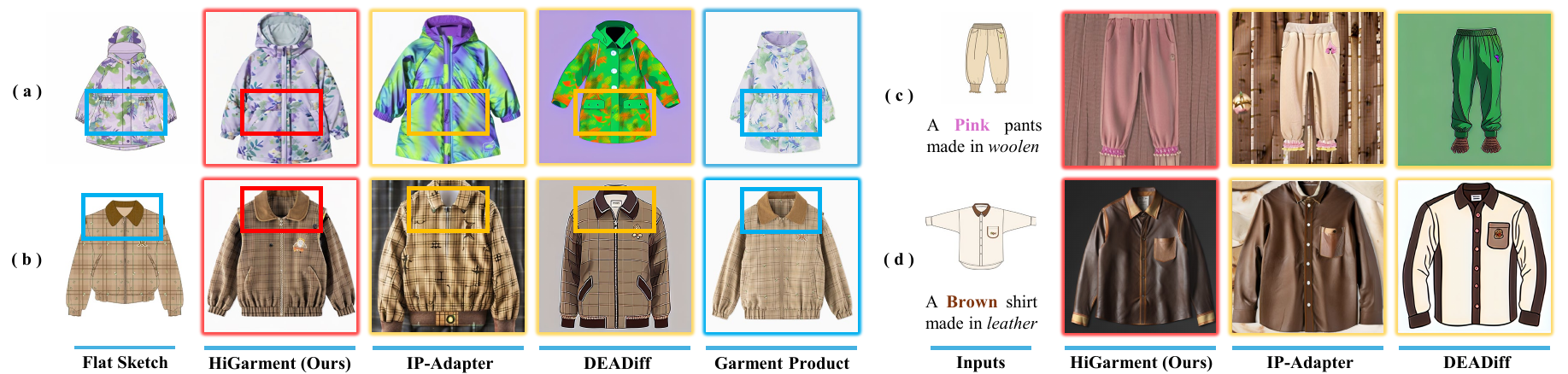}\\[1ex]
  \end{center}
  \begin{justify}
    \small
    \refstepcounter{figure}
    Figure~\thefigure: Flat Sketch to Realistic Garment task generates a realistic sample garment from a flat sketch and text prompts, providing a reference for factories to produce garment product. HiGarment captures the structure of garments and attributes (color and fabric) details. \textcolor{blue}{Blue} boxes represent target results, \textcolor{red}{red} boxes describe accurate results, while \textcolor{orange}{yellow} boxes describe inaccurate results.
  \label{fig:two}
  \end{justify}
  \bigskip
}
\maketitle
\renewcommand{\thefootnote}{\fnsymbol{footnote}}
\footnotetext[2]{Equal contribution.}
\footnotetext[1]{Corresponding author.}
\makeatother

\maketitle
\begin{abstract}
Diffusion-based garment synthesis tasks primarily focus on the design phase in the fashion domain, while the garment production process remains largely underexplored. To bridge this gap, we introduce a new task: Flat Sketch to Realistic Garment Image (FS2RG), which generates realistic garment images by integrating flat sketches and textual guidance. FS2RG presents two key challenges: 1) fabric characteristics are solely guided by textual prompts, providing insufficient visual supervision for diffusion-based models, which limits their ability to capture fine-grained fabric details; 2) flat sketches and textual guidance may provide conflicting information, requiring the model to selectively preserve or modify garment attributes while maintaining structural coherence. To tackle this task, we propose HiGarment, a novel framework that comprises two core components: i) a multi-modal semantic enhancement mechanism that enhances fabric representation across textual and visual modalities, and ii) a harmonized cross-attention mechanism that dynamically balances information from flat sketches and text prompts, allowing controllable synthesis by generating either sketch-aligned (image-biased) or text-guided (text-biased) outputs. Furthermore, we collect Multi-modal Detailed Garment, the largest open-source dataset for garment generation. Experimental results and user studies demonstrate the effectiveness of HiGarment in garment synthesis. The code and dataset are available at https://github.com/Maple498/HiGarment. 
\end{abstract}

\section{Introduction}
\label{sec:intro}
In garment production, a key step involves converting flat sketches, a detailed two-dimensional design outlining garment details \cite{wan2014shape}, into physical samples through multiple stages, including fabric sourcing, pattern creation, and detail incorporation. However, this process causes prolonged material sourcing and repeated iterations for localized modifications. For instance, adjustments such as adding pockets or modifying collars require frequent flat sketch revisions and sample reproductions. These inefficiencies significantly extend production timelines and resource consumption, making traditional workflows inadequate for the fast-paced demands of the modern fashion industry \cite{aus2021designing}.

Previous garment synthesis tasks have primarily focused on the early creative stages of design but failed to meet garment production demands. Text-to-garment synthesis \cite{zhang2023diffcloth, zhang2024garmentaligner} generates images from descriptive text but lacks structural precision due to the absence of visual references, failing to generate images that accurately reflect production outcomes that conform to flat sketches. Sketch-driven garment synthesis \cite{zhang2022armani, chen2023foldgen, zhang2024texcontrol} ensures structural accuracy using initial sketches but cannot support precise fabric and detail editing using textual input, leading to such approaches incompatible with garment production pipelines. Therefore, generating realistic garment images with support for flexible attributes and localized element modification holds significant potential to streamline garment production but remains under-explored.



In this paper, we propose a new task, Flat Sketch to Realistic Garment Image (FS2RG), which aims to generate photorealistic garment images from flat sketches and text prompts. Realistic garment images, as defined in our work, ensure alignment with the flat sketches in structure, color, pattern, and photorealistic sewing visual effects. Compared with previous garment synthesis tasks, FS2RG addresses two critical but practical requirements: 1) to preserve precise visual alignment with flat sketches in layout, patterns, and color while enabling realistic fabric representation through textual guidance, and 2) to allow designers to adjust garment attributes during the sample creation process using simple textual inputs.
In this task, the flat sketch provides rich visual information about the garment, while text serves as the primary guidance to ensure ease of garment production. Therefore, these two modalities offer distinct yet collaborative types of information. This task requires the image generation model to achieve the complementary relationship between the image and text modalities. We define this complementary relationship as \textbf{modality harmony}. Traditional diffusion-based generation methods face two main challenges in the FS2RG task. \textbf{First}, these methods are designed to align image and text features under the assumption that both modalities provide corresponding information. This mismatch causes the fabric attribute (only guided by text modality in the FS2RG task) to be often overlooked during the alignment process, as the model is biased towards matching features that appear in both modalities. As shown in Fig. \hyperref[fig:two]{1}, the fabric in (a) and the collar color in (b) of the images generated by IP-Adapter \cite{ye2023ip} are not accurately reproduced. \textbf{Second}, these methods lack the ability to actively select between modalities when inconsistencies arise. As a result, when user instructions do not match the image features, the generated output fails to accurately reflect the intended modifications. The pant in Fig. \hyperref[fig:two]{1} (c) generated by IP-Adapter \cite{ye2023ip} retains the same color as the reference image, unable to accurately reflect the color to pink based on the text prompt.

Inspired by the above analysis, we introduce HiGarment, a cross-modal \textbf{H}armony-based d\textbf{i}ffusion model for Flat Sketch to Realistic \textbf{Garment} task. HiGarment aims to achieve modality harmony by accurately translating flat sketches into realistic garment images. It contains a multi-modal semantic enhancement mechanism to improve fabric representation in visual and textual features by retrieving fabric samples from an extra vector database. This module injects fine-grained fabric features that cannot be captured by simple feature aggregation from textual and visual modalities. Additionally, we develop a novel harmonized cross-attention mechanism to guide the generative model in selectively favoring either text or image guidance during generation. Unlike previous approaches \cite{ye2023ip, mou2024t2i}, this mechanism dynamically balances and resolves conflicts between modalities, enabling precise control over garment attributes and structure in the generated images.

The current largest fashion dataset, CM Fashion \cite{zhang2022armani}, is not open source up to now, and DressCode \cite{he2024dresscode} has insufficient textual content. Given the lack of professional garment dataset, we collect \textbf{M}ulti-\textbf{M}odal \textbf{D}etailed \textbf{Garment} (MMDGarment), a manually annotated multi-modal garment dataset, which contains 1) 20,151 garment images with detailed close-up images and flat sketches; 2) corresponding manually annotated text labels covering 11 fabric types, 23 color categories, and localized detail description; 3) a vector database that contains 150 fabric types with corresponding sample fabric image to enhance the control ability of garment fabric.
The contributions of this paper are summarized as follows:
\begin{itemize}
\item We present Flat Sketch to Realistic Garment Image, a new task in the fashion area focused on the garment production process. This task aims to streamline the process of producing physical garment prototypes by generating realistic 2D images. 

\item We collect the largest open-sourced multi-modal fashion dataset, MMDGarment, containing garment images with corresponding text annotations, flat sketches, close-up images, and fabric vector database.

\item We introduce a new method, HiGarment, to tackle the FS2RG task. HiGarment emphasizes the importance of harmonizing visual and textual modalities by capturing their semantics and balancing them adaptively. Qualitative and quantitative results show that our method achieves state-of-the-art performance in the FS2RG task.


\end{itemize}


\begin{figure*}[htbp]
  \centering
  \includegraphics[width=1\textwidth]{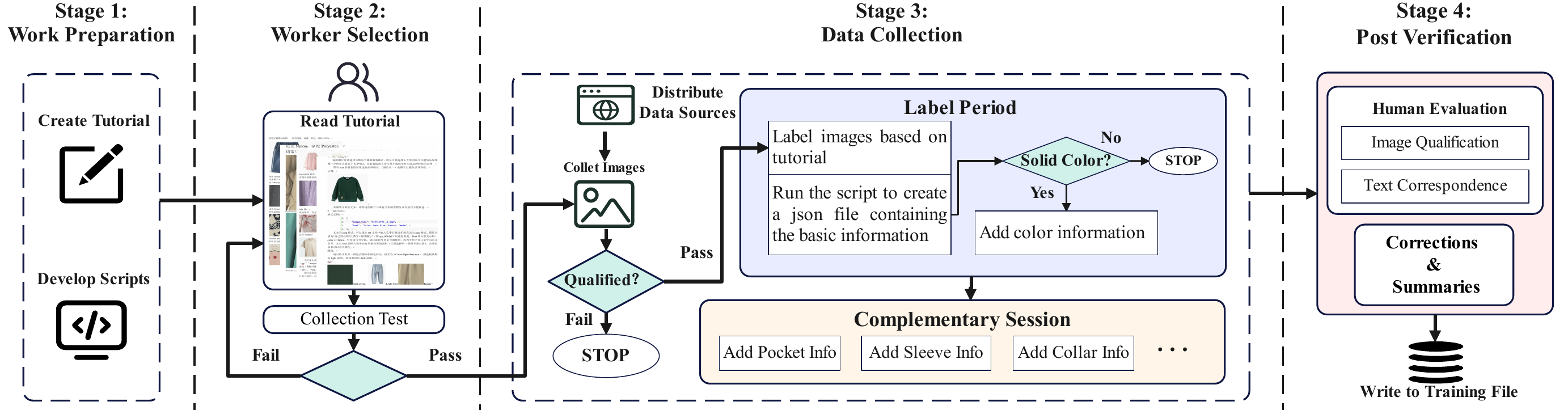}  
  \caption{The pipeline of MMDGarment dataset collection.} 
  \label{fig:four}
\end{figure*}

\section{Related Works}

\subsection{Multi-modal garment synthesis}
Previous garment synthesis tasks primarily focus on early design stages, categorized into two main approaches: 1) text-to-garment synthesis \cite{zhang2023diffcloth, zhang2024garmentaligner}, which generates garment images from descriptive text but fails to incorporate visual references. Some methods focus on aligning garment attributes and structure using text input. For example, Diffcloth \cite{zhang2023diffcloth} and GarmentAligner \cite{zhang2024garmentaligner} use segmented captions and semantic garment segmentation to achieve component-level alignment. However, these approaches are limited to extracting garment features only from textual descriptions and cannot interpret visual inputs. 2) Sketch-driven garment synthesis \cite{zhang2022armani, chen2023foldgen, zhang2024texcontrol} relies on initial design sketches as structural guides for image generation, leading the model to prioritize structure while missing finer details, such as textual attributes and visual nuances. However, in this task \cite{zhang2022armani, chen2023foldgen}, sketches serve only as structural guides; hence, the sketches used in these methods are not well-suited for the fashion design process, limiting their practicality in the garment industry. In this paper, we investigate a new multi-modal garment synthesis task that generates realistic garment images using sketch-based structural input combined with text-based attribute descriptions.

\subsection{Diffusion-based image generation}

Recently, Diffusion Probabilistic Models \cite{sohl2015deep} have excelled in image-generation tasks \cite{dhariwal2021diffusion, ho2020denoising} by recovering target data distributions disrupted during the forward diffusion process.
Latest diffusion models \cite{feng2024enhancing, liu2024residual, sauer2025adversarial} leverage large-scale pre-training and U-Net \cite{kawano2024maskdiffusion,ma2024u,lin2022ds} to integrate text features from pre-trained encoders \cite{li2024blip,radford2021learning,gupta2023cliptrans,liu2023ed}. Recent works in controllable image generation \cite{zhang2023adding, mou2024t2i, zhao2024uni, ye2023ip}, based on Stable Diffusion, enhance control while avoiding the laborious work of fine-tuning large models. ControlNet \cite{zhang2023adding} demonstrates that an adapter could be trained with a pre-trained text-to-image diffusion model to learn specific input conditions. To reduce the fine-tuning cost, Uni-ControlNet \cite{zhao2024uni} presents a multi-scale condition injection strategy to learn an adapter for various local controls. Meanwhile, IP-adapter \cite{ye2023ip} employs a decoupled cross-attention module to align text and image prompts in the denoising process. Inspired by the effectiveness of lightweight adapter, we propose a new approach that further achieves complementarity between structural features of flat sketches and attribute-level prompt (e.g., color, fabric) for the FS2RG task.

\begin{figure*}[htbp]
  \centering
  \includegraphics[width=1\linewidth]{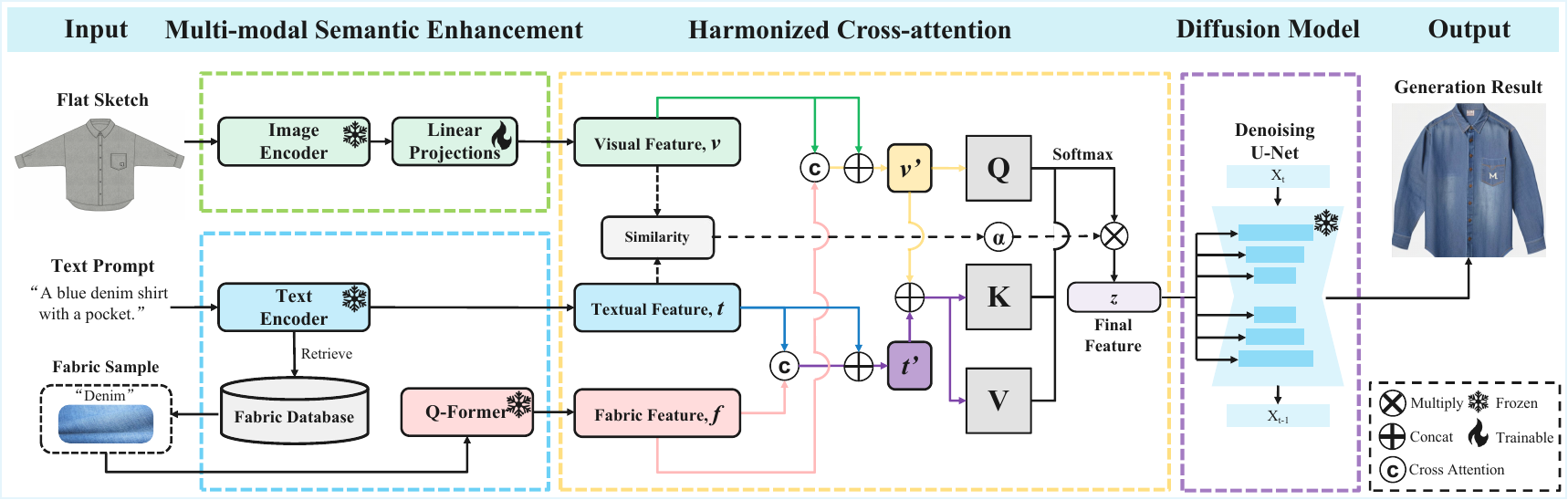}  
  \caption{The overall framework of HiGarment. We design a multi-modal semantic enhancement mechanism to extract representations separately from images and text, along with a harmonized cross-attention mechanism to selectively align image and text representations for modality harmony.} 
  \label{fig:three}
\end{figure*}

\subsection{Fashion datasets}
Fashion datasets are generally divided into try-on datasets \cite{morelli2022dress} and garment synthesis datasets \cite{han2018viton, zhang2022armani}. DressCode \cite{morelli2022dress} is a virtual try-on dataset that contains over 50,000 pairs of garment images and human poses. However, its text annotation is designed for virtual try-on tasks, which are unsuitable for garment generation tasks.
For garment synthesis datasets, VITON \cite{han2018viton} includes 32,506 images and annotations, but it lacks localized annotations for garment elements and fabric annotations. CM-Fashion \cite{zhang2022armani} contains 500,000 garment images and corresponding text descriptions, but it is not a publicly available dataset. Therefore, the demand for an open-sourced multi-modal fashion dataset with detailed garment attribute annotations is necessary.

\section{Multi-Modal Detailed Garment Dataset}
We build a multi-modal fashion dataset named Multi-Modal Detailed Garment (MMDGarment), specially designed for the FS2RG task. To the best of our knowledge, this is the first publicly available fashion dataset that includes garment attributes, detailed annotations of garment components, and corresponding flat sketches.
\subsection{Dataset collection} 
The dataset collection team followed standardized protocols to ensure high consistency and quality. All images were annotated with detailed garment attributes, and annotation accuracy was verified in consultation with professional garment designers. The dataset also features technical flat sketches created by designers from our collaborating company. An overview of the collection pipeline is shown in Fig. \ref{fig:four}, and further details are provided in Supp. \ref{sec:dataset}.



\subsection{Dataset statistics and analysis}
The MMDGarment dataset contains 20,151 garment images with 11 fabric types, 23 color categories, and localized detail descriptions. Each text annotation includes details about the garment's color, fabric, pattern, structure, and style. In addition, MMDGarment provides a database of 150 fabric image-text pairs, further enriching the variety of fabric references for fabric representation. Notably, the test set of the MMDGarment dataset contains 1,000 text-image pairs, including 845 flat sketches paired with photos of corresponding produced garments, enabling detailed comparisons and providing a reliable benchmark for evaluation. More details about dataset can be found in the Supp. \ref{sec:dataset}.


\section{Method}
\subsection{Task definition}
The FS2RG task aims to generate realistic garment images \(R\) by integrating complementary information from sketch image \(I\) and text inputs \(T\). The \(I\) is converted to visual feature $v \in \mathbb{R}^{d}$, and \(T\) is converted to textual feature $t \in \mathbb{R}^{d}$, in dimension \(d\). The goal of this task is to leverage these complementary features to achieve modality harmony. The output multi-modal embedding is represented as  $z \in \mathbb{R}^{d}$. Finally, \(z\) is provided as input to the U-Net in Stable Diffusion to generate the realistic garment image \(R\). 

\subsection{Model architecture}
\subsubsection{Overview}
We propose HiGarment to address the FS2RG task. As shown in \cref{fig:three}, HiGarment contains two parts: 1) Multi-modal semantic enhancement mechanism processes visual concepts and attribute information from the flat sketch and text prompts, enabling the model to learn semantic information across different modalities effectively; 2) Harmonized cross-attention mechanism achieves cross-modal harmony by dynamically selecting the desired visual and textual representation from Multi-modal semantic enhancement mechanism. It enables the model to dynamically adjust the output to align more closely with either the image or the text.

\subsubsection{Multi-modal semantic enhancement mechanism}
To fully leverage information from both visual and textual modalities, we introduce the multi-modal semantic enhancement mechanism (MMSE). While directly using CLIP \cite{radford2021learning} as the feature extraction approach, it presents two limitations: 1) the diversity of fabrics makes it challenging to describe them accurately with text alone, leading to generated garment images that fail to fully capture the intended fabric details. Traditional diffusion-based methods rely on prior knowledge in CLIP when handling garment fabrics, leading to inaccuracies when dealing with uncommon fabric materials; 2) existing feature extraction methods utilize image features and text features to achieve global semantic alignment but lack explicit modeling of fine-grained attributes (e.g., fabric texture in FS2RG tasks), leading to insufficient control over local details. In this work, we adopt the multi-modal semantic enhancement mechanism to improve the ability of textual and visual modalities to process fabric attributes.

To capture the high-level visual semantics from the flat sketch, we train a linear projection using the MMDGarment to automatically extract garment visual features \(v\) containing color, layout, etc., from reference images \(I\) without relying on additional tools like masks or segmentors \cite{zhang2023diffcloth}, meeting the convenience requirements of the garment production process. The textual features \(v\) from the text prompts \(T\) are directly extracted through the CLIP text encoder. 

\begin{figure}[h]
  \centering
  \includegraphics[width=0.85\linewidth]{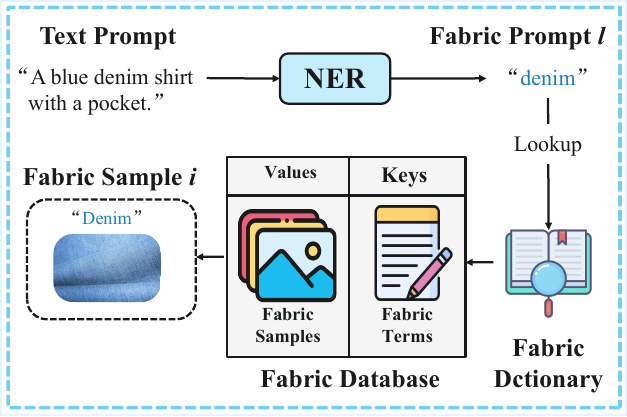}  
  \caption{Illustration of the retrieval process in the fabric database.} 
  \label{fig:database}
\end{figure}

To supplement the visual information of the fabric features, we construct a scalable fabric sample database to provide additional visual information for fabric attributes. The fabric text embeddings are stored as keys, and their corresponding sample images as values. Specifically, we first extract the fabric label $l$ from the text prompt $t$ using Named Entity Recognition (NER), and then standardize it using a predefined fabric dictionary. This dictionary maps variant fabric terms (e.g., jeans) to a normalized term (e.g., denim). The normalized fabric term is then used to retrieve the corresponding fabric name from our database. An illustration of this process is shown in Fig. \ref{fig:database}. Finally, we obtain the associated image $i$ for downstream use. 


Inspired by Q-Former's \cite{li2023blip} ability to establish deep semantic connections between visual and textual modalities, MMSE leverages contextual information from fabric samples to enhance the understanding of the textual description. The fabric sample \(i\) and the fabric label \(l\) are processed through Q-Former, which aligns and integrates complementary information from both modalities. Specifically, the queries \(q\) act as a fabric dictionary indexed by sample \(i\) and label \(l\). The self-attention mechanism retrieves target-specific semantics from this dictionary:
\begin{equation}
f_{s}=\text{Softmax}(\frac{[q;l]W_{q}^{s}([q;l]W_{k}^{s})^{T}}{\sqrt{d}})[q;l]W_{v}^{s}
\label{eq:qf1}
\end{equation}
whereas the cross-attention integrates label and sample image features to derive the optimal representation:
\begin{equation}
f=\text{Softmax}(\frac{f_{s}W_{q}^{c}(iW_{k}^{c})^{T}}{\sqrt{d}})iW_{v}^{c}
\label{eq:qf2}
\end{equation}
where \(W_{q}^{s}\), \(W_{k}^{s}\), \(W_{v}^{s}\) are the projection matrices from self-attention block and \(W_{q}^{c}\), \(W_{k}^{c}\), \(W_{v}^{c}\) are the projection matrices from cross-attention block. 

Finally, HiGarment enhances the visual and textual embeddings by incorporating fabric information through cross-attention. For the visual feature enhancement, we define the Query as \(Q_v = vW^Q\) and use the concatenation of the text and fabric features, \(t \oplus f\), to form the Key \(K_{tf}\) and Value \(V_{tf}\). The enhanced visual feature is then given by:
\begin{equation}
v' = v \oplus \text{Attention}(Q_v,\, K_{tf},\, V_{tf}).
\label{eq:enhancev}
\end{equation}
Similarly, for the text feature enhancement, we set \(Q_t = tW^Q\) and use \(v \oplus f\) to obtain \(K_{vf}\) and \(V_{vf}\), resulting in an enhanced text feature: 
\begin{equation}
t' = t \oplus \text{Attention}(Q_t,\, K_{vf},\, V_{vf}).
\label{eq:enhancet}
\end{equation}
This process enhances the semantic representation of the text by incorporating visual cues from the fabric sample, enabling precise control over fabric attributes. Finally, MMSE outputs the visual feature \(v'\) and textual feature \(t'\), which are subsequently used in the further modality harmonization stage.

\subsubsection{Harmonized cross-attention mechanism}
The FS2RG task necessitates the model to conform to the image's visual effect while following text guidance during the generation process when modality inconsistencies arise. The original cross-attention mechanisms fail to prioritize between modalities when textual and visual modalities are inconsistent, causing the generative model to focus solely on aligned features. It becomes challenging to support the frequent modifications in the FS2RG task. We propose a novel harmonized cross-attention mechanism (HCA) to dynamically harmonize the complementary information from both modalities by adapting to the unique contributions of visual and textual information. The harmonized cross-attention further guides the model in determining image-biased or text-biased outputs during the generation process.

The harmonized cross-attention mechanism first evaluates differences between \( v \) and \( t \) from the MMSE. We calculate the cosine similarity \( s \) between \( v \) and \( t \) as follows:
\begin{equation}
\
s = \frac{v \cdot t}{\| v \| \| t \|}
\
\label{qu:3}
\end{equation}
where \( s \) with higher values indicating stronger alignment between the image and the text description, meaning the generation process tends to visual modality, and vice versa. The similarity \( s \) is defined in the range \((-1, 1)\). To modulate the influence of the dynamic selection mechanism on the image generation process, we calculate a weight factor \( \alpha_{\text{}} \) ranged \((0, 1)\) using the Sigmoid function as follows:

\begin{equation}
\alpha(s) = \lambda + (1 - \lambda) \cdot \sigma(s)
\label{eq:4}
\end{equation}
where $\sigma(s) = \frac{1}{1 + e^{-s}}$. This weight factor is designed as a conditioning parameter for the harmonized cross-attention mechanism. Higher similarity drives the model toward image-biased generation, closely reflecting the flat sketch. In comparison, lower similarity shifts the focus toward text-biased generation, ensuring the output more accurately embodies the textual input. We will analyze in section 5.5 to explain the value of parameter \(\lambda\) in the Sigmoid function. 

The harmonized cross-attention process can be formalized as follows:  
\begin{equation} 
Q = v' W^Q
\label{eq:query}
\end{equation}  
\begin{equation} 
K = \text{Concat}(v' W^K, t' W^K)
\label{eq:key}
\end{equation}  
\begin{equation} 
V = \text{Concat}(v' W^V, t' W^V)
\label{eq:value}
\end{equation}  
Here, \(W^Q\), \(W^K\), and \(W^V\) are projection matrices that map the enhanced image embedding \(v'\) and the enhanced text embedding \(t'\) into the query, key, and value spaces, respectively. The query feature \(z\) is then computed by adjusting the standard attention formula using the dynamic weight factor \( \alpha \), derived from the Eq. \ref{eq:4}, as follows:  
\begin{equation}
    z = \alpha \cdot \text{Softmax}\!\Bigl(\tfrac{Q K^T}{\sqrt{d}}\Bigr)\,V
\label{eq:attention}
\end{equation}  
finally, \(z\) is integrated into the pre-trained Stable Diffusion UNet by injecting it into each cross-attention layer. Consequently, a larger $\alpha$ increases the influence of the external image condition, while a smaller $\alpha$ reduces this influence and makes the model rely more on text guidance. Unlike the standard diffusion model, which directly combines image and text features, the harmonized cross-attention mechanism ensures seamless modality harmony by aligning and enhancing the representation of garment attributes and structures, leading to high-quality generated images.

\section{Experiment}

\subsection{Experimental setup}
\textbf{Implementation details.} We adopt Stable Diffusion v1.5 \cite{rombach2022high} as the base model. The OpenCLIP ViT-H/14 \cite{radford2021learning} is used as the image encoder, while the CLIP text encoder processes the text input. The MMSE contains 85.15M trainable parameters. All training data is resized to 512 × 512 for processing. During the training process, the AdamW \cite{loshchilov2017decoupled} optimizer is set with a learning rate of 0.0001 and weight decay of 0.01 with a batch size of 8. We utilize mean squared error as the loss function. The training process is completed with L40 GPUs for 30 hours. In the inference stage, we set the DDIM sampler with 50 steps. 


\noindent\textbf{Baselines and evaluation metrics.} 
Since FS2RG is a new task, there are no existing baselines. For comprehensiveness, we compare the performance of HiGarment on the MMDGarment dataset to the state-of-the-art methods. We select two types of methods: 1) global image editing: DEADiff \cite{qi2024deadiff}; and 2) multi-modal prompts-based image generation: IP-Adapter \cite{ye2023ip}, Uni-Controlnet \cite{zhao2024uni}, Versatile Diffusion \cite{xu2023versatile}, and SSR-Encoder \cite{zhang2024ssr}. 
To ensure fair comparisons, all open-source models were trained on the MMDGarment dataset, and the official inference code from each baseline was used for evaluation.
We employ CLIPScore \cite{hessel2021clipscore}, FID \cite{heusel2017gans}, and LPIPS \cite{zhang2018unreasonable} to evaluate the quality of the generation results. Specifically, the evaluation measures the similarity between generated images and product images in the MMDGarment test set. Additionally, we conducted a user study and employed a multi-modal large language model evaluation to further demonstrate the effectiveness of HiGarment.




\subsection{Qualitative results}
\begin{figure*}[htbp]
\centering
\includegraphics[width=0.95\textwidth]{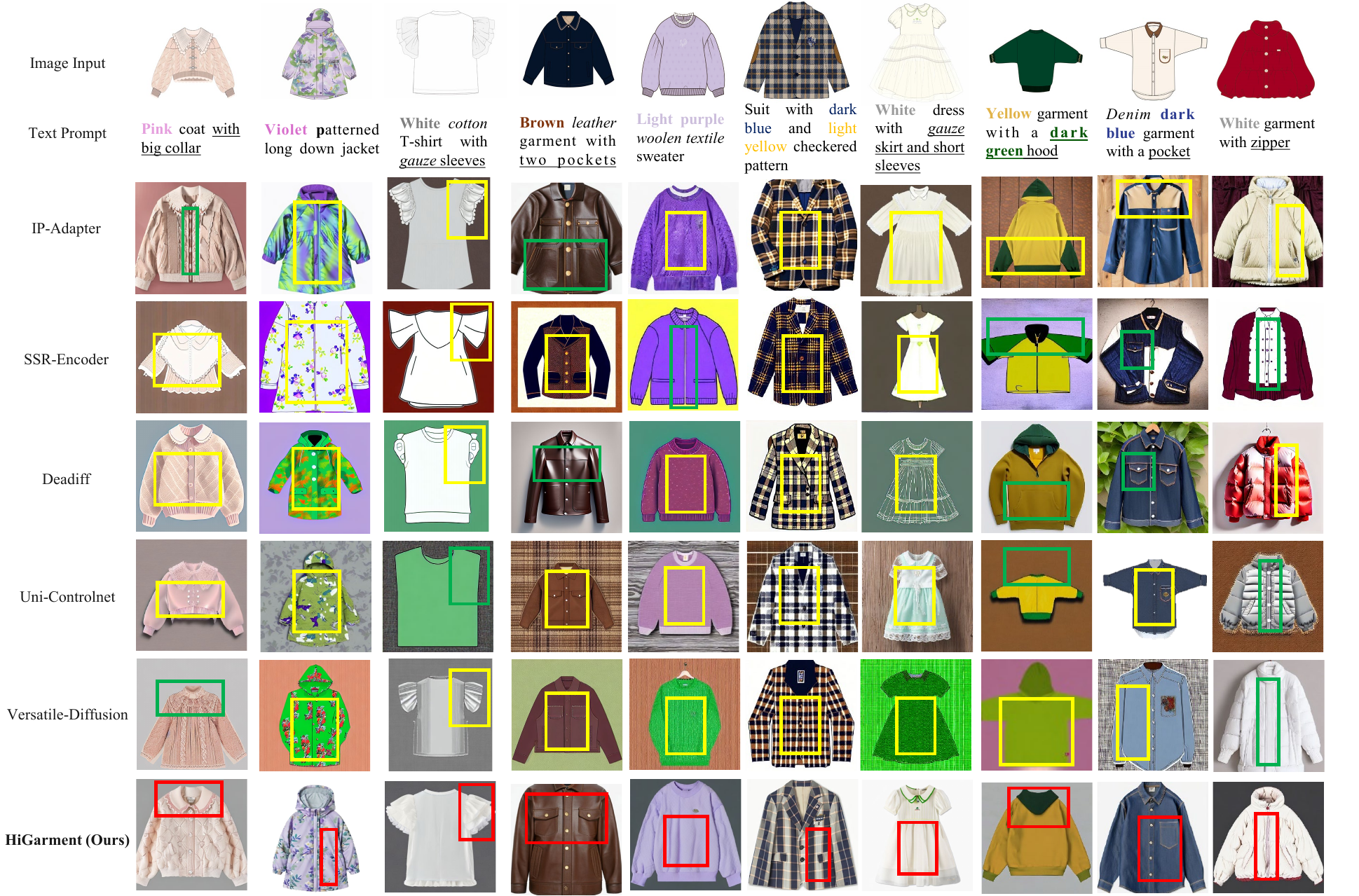}
\caption{Qualitative comparison of real garment image generation ability between our method and others. We use \textbf{colorful words} to represent color; \textit{italics} to represent fabric; and \underline{underline} to represent the garment component. \textcolor{red}{Red} boxes indicate accurate results, \textcolor{myYellow}{yellow} boxes indicate incorrect fabric, and \textcolor{myGreen}{green} boxes indicate incorrect structure.}\label{fig:five}
\end{figure*}

\noindent\textbf{Main results.} 
Fig.~\ref{fig:five} shows the qualitative comparison results. Since a flat sketch is a unique reference type, existing methods show limited performance (shown in the yellow boxes). For instance, DEADiff~\cite{qi2024deadiff} preserves garment patterns and colors but fails to accurately capture structural and fabric details. Similarly, while Uni-ControlNet~\cite{zhao2024uni} maintains garment structures consistent with the flat sketches, the resulting images inadequately represent key attributes, leading to unrealistic outputs. In contrast, HiGarment accurately reproduces garment structure, patterns, and fabric, creating more realistic garment images.

Our method also demonstrates the ability to modify garment attributes globally (e.g., changing colors) and locally (e.g., adding components such as hoods or pockets). Furthermore, HiGarment exhibits high sensitivity to material and fabric details, accurately generating textures like polyester and denim. The third column in Fig. \ref{fig:five} demonstrates the case of a single garment with different fabrics. HiGarment shows the capability to distinguish different fabrics. Comparison methods show limited capability in altering garment structures and attributes, underscoring the superiority of HiGarment in such transformations.

In addition, we evaluated cases where the text input did not specify any fabric or included fabric names not present in the database. If the fabric is missing, the model defaults to generating a fabric-like result based on the visual context. If the fabric label is not found in the database, we query our predefined fabric dictionary with the user’s input and select the most semantically similar fabric entry during retrieval.

\noindent\textbf{Qualitative discussion.} Our analysis identifies two main reasons for the above observations: \textbf{First}, style transfer approaches~\cite{qi2024deadiff} that prioritize style or visual attributes focus primarily on appearance during training, providing insufficient guidance on garment-related elements such as layout or components. This shortcoming leads to outputs that fail to align with the appearance of real garments. \textbf{Second}, multi-modal prompt methods~\cite{zhao2024uni, ye2023ip} can generate images aligning closely with the structure of the flat sketch but often fail to accurately capture critical garment attributes (e.g., fabric textures). Their emphasis on general visual-textual alignment hinders the integration of fine-grained attribute information conveyed by the text. This highlights the importance of modality harmony in the FS2RG task for producing realistic garment images.

\begin{table}[ht]
\centering
\scalebox{0.6}{
\begin{tabular}{lccc}
\toprule
 & CLIPScore ($\uparrow$)  & FID ($\downarrow$) & LPIPS ($\downarrow$) \\
\midrule
IP-Adapter \cite{ye2023ip}          & \textcolor{definedBlue}{0.7768} & \textcolor{definedBlue}{11.0334} & 0.7025 \\
Uni-Controlnet \cite{zhao2024uni}   & 0.6869                         & 16.9257                         & 0.8369 \\
Versatile Diffusion \cite{xu2023versatile} & 0.6639                    & 13.6090                         & 0.7979 \\
DEADiff \cite{qi2024deadiff}        & 0.6459                         & 14.5238                         & \textcolor{definedBlue}{0.6780} \\
SSR-Encoder \cite{zhang2024ssr}     & 0.6220                         & 16.4208                         & 0.7425 \\
\textbf{HiGarment (Ours)}           & \textcolor{definedPink}{\textbf{0.8054}} & \textcolor{definedPink}{\textbf{8.8109}} & \textcolor{definedPink}{\textbf{0.6008}} \\
\bottomrule
\end{tabular}}
\caption{Quantitative comparisons between our method and others. \textcolor{definedPink}{Pink} and \textcolor{definedBlue}{blue} highlight the best and second-best results for each metric, respectively.}
\label{table1}
\end{table}

\subsection{Quantitative results}
As shown in Tab. \ref{table1}, HiGarment outperforms the state-of-the-art method \cite{ye2023ip} by 3.7\%, 20.3\%, and 14.5\% for CLIPScore, FID, and LPIPS, respectively. We draw several conclusions from this table. \textbf{First}, compared with \cite{zhao2024uni}, which uses flat sketch as a structure condition, HiGarment significantly outperforms in all evaluation matrices. This demonstrates the specialization of flat sketches. Compared with \cite{ye2023ip}, which learns sufficient information from images, the result shows the effectiveness of applying a garment-specific feature extraction mechanism (e.g., MMSE) and harmonizing text and image modalities (e.g., HCA). \textbf{Second}, DEADiff \cite{qi2024deadiff} performs well in LPIPS but scores lower in CLIPScore and FID. We attribute this to the fact that LPIPS simulates human visual perception of images, while CLIPScore and FID focus more on feature-level similarity. These findings demonstrate the limitations of existing methods in the fashion domain, especially in the FS2RG task. 

\begin{table}[ht]
\centering
\scalebox{1}{
\resizebox{\linewidth}{!}{
\begin{tabular}{lccc}
\toprule
                 & MLLM Evaluation ($\uparrow$) & Experts Evaluation ($\uparrow$) & Non‑experts Evaluation ($\uparrow$) \\
\midrule
IP‑Adapter \cite{ye2023ip}          & \textcolor{definedBlue}{6.355} & \textcolor{definedBlue}{6.958} & \textcolor{definedBlue}{6.244} \\
DEADiff \cite{qi2024deadiff}        & 5.698                           & 4.799                           & 5.542                           \\
\textbf{HiGarment (Ours)}           & \textcolor{definedPink}{\textbf{6.996}} & \textcolor{definedPink}{\textbf{7.735}} & \textcolor{definedPink}{\textbf{6.853}} \\
\bottomrule
\end{tabular}
}}
\caption{Comparison between human‑evaluation results and MLLM evaluation results. \textcolor{definedPink}{Pink} and \textcolor{definedBlue}{blue} highlight the best and second‑best results for each metric, respectively.}
\label{table2}
\end{table}

\subsection{MLLM-based evaluation and user study}
Inspired by \cite{chen2024mllm}, we use a Multi-modal Large Language Model ChatGPT 4o\footnote{\url{https://openai.com/chatgpt}} to rate the generation results across four aspects: structure, color, fabric, and details. We also invited 2 professional garment designers, 11 fashion design students, and 20 non-experts to assess the similarity between generated and real garment images. As shown in Tab.~\ref{table2}, HiGarment achieves the highest scores in both, highlighting its performance. More details about MLLM-based evaluation and user study can be found in the Supp.~\ref{sec:mllm}.

\subsection{Ablation study and parameter analysis}
\begin{figure}[ht]
    \centering
    \begin{subfigure}[b]{0.38\textwidth}
        \centering
        \includegraphics[width=\textwidth]{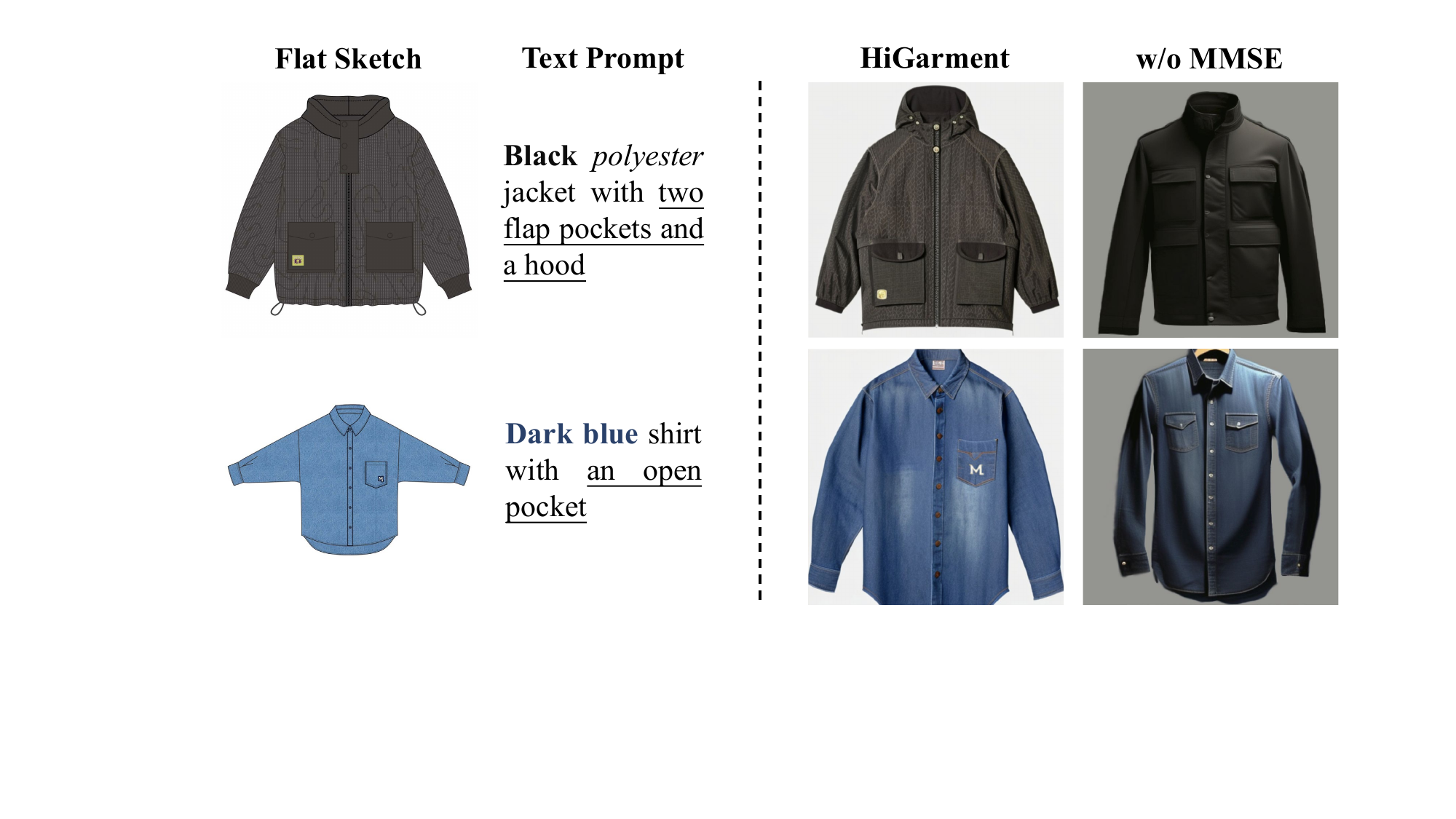}
        \caption{Comparisons between original HiGarment and HiGarment without MMSE.}
        \label{fig:image_a}
    \end{subfigure}
    \hfill
    \begin{subfigure}[b]{0.38\textwidth}
        \centering
        \includegraphics[width=\textwidth]{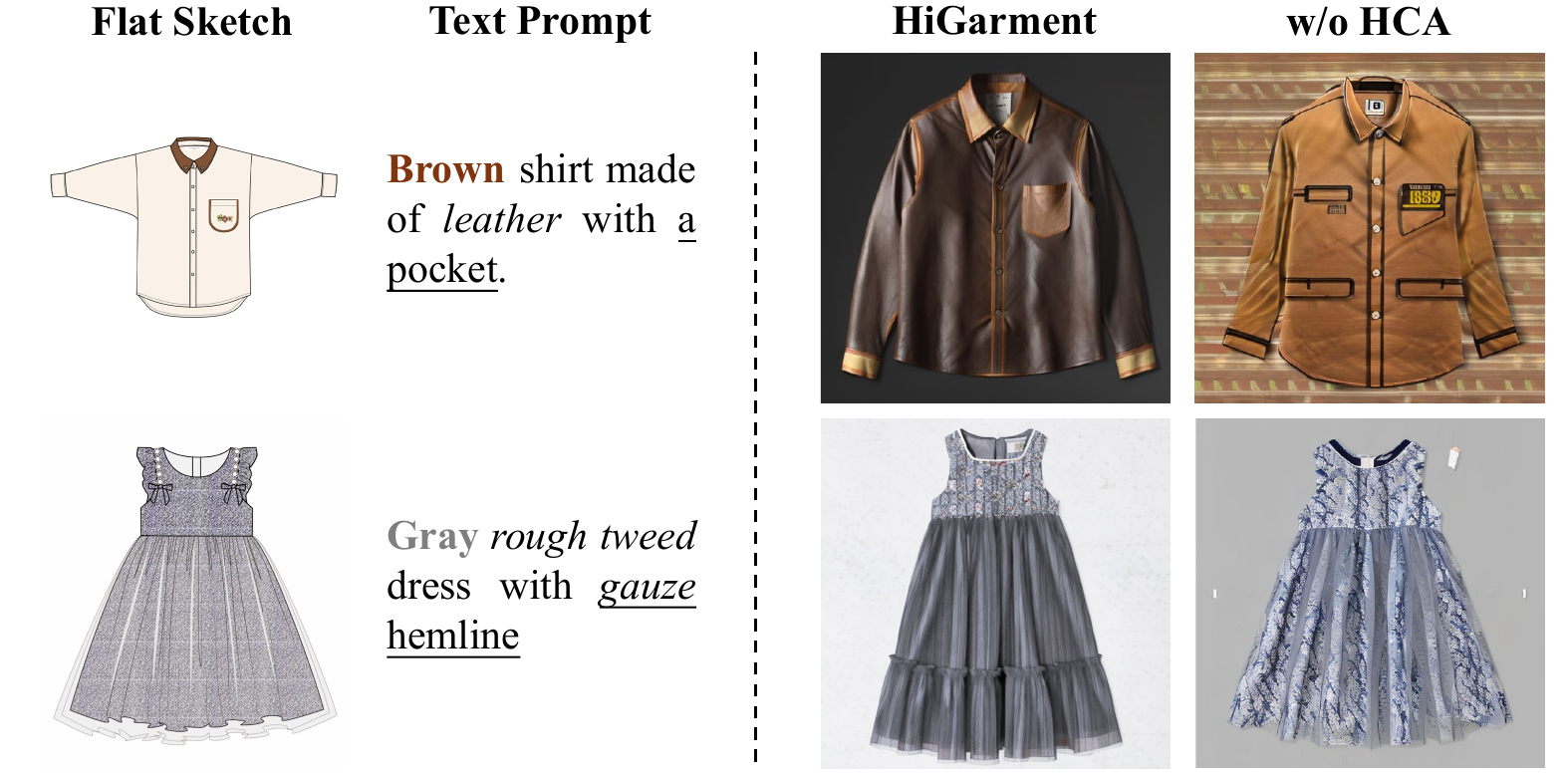}
        \caption{Comparisons between original HiGarment and HiGarment without harmonized cross-attention.}
        \label{fig:image_b}
    \end{subfigure}
    \caption{Ablation study results.}
    \label{fig:example_subcaption}
\end{figure}

\textbf{Ablation study.} 
We conducted ablation studies to assess the significance of the key modules in HiGarment. We replace the MMSE with the pre-trained feature extractor from SD1.5; the embeddings from the image and text encoders are directly passed to the UNet of the diffusion model. As shown in Tab. \ref{table:ab}, the removal of the MMSE module led to a decline in CLIPScore \cite{hessel2021clipscore}, FID \cite{heusel2017gans}, and LPIPS \cite{zhang2018unreasonable} by 9.70\%, 36.15\%, and 11.35\%, respectively. 
Fig. \hyperref[fig:example_subcaption]{6 (a)} illustrates that without MMSE, HiGarment fails to render components and fabric attributes from the flat sketch accurately. This indicates that MMSE provides critical garment-specific representation, especially for attributes across visual and textual modalities that are difficult to integrate. Besides, removing the HCA resulted in a more significant performance drop, with CLIPScore, FID, and LPIPS depreciating by 15.16\%, 23.71\%, and 21.29\%, respectively. In this scenario, the embeddings extracted by MMSE were directly concatenated instead of being processed through HCA. As shown in Fig. \hyperref[fig:example_subcaption]{6 (b)}, the absence of HCA causes HiGarment to fail in effectively harmonizing visual and textual references, leading to suboptimal integration of attribute features. These results underline the critical roles of MMSE and HCA in generating images with precise attribute rendering and modality harmony. Besides, we fine-tuned HiGarment on VITON-HD~\cite{han2018viton} to disentangle the performance gains contributed by the proposed MMDGarment dataset from those of the HiGarment framework itself. The results demonstrate that the improvements are primarily driven by our designed modules, rather than by dataset scale alone. 

\begin{table}[ht]
\centering
\scalebox{0.65}{
\begin{tabular}{lccc}
\toprule
 & CLIPScore~($\uparrow$) & FID~($\downarrow$) & LPIPS~($\downarrow$) \\
\midrule
w/o MMSE       & 0.7342 & 13.7995 & 0.6777 \\
w/o HCA        & 0.6994 & 11.5491 & 0.7633 \\
w/o MMDGarment & \textcolor{definedBlue}{0.7635} & \textcolor{definedBlue}{9.4932} & \textcolor{definedBlue}{0.6385} \\
\textbf{HiGarment}
               & \textcolor{definedPink}{\textbf{0.8054}} & \textcolor{definedPink}{\textbf{8.8109}} & \textcolor{definedPink}{\textbf{0.6008}} \\
\bottomrule
\end{tabular}
}
\caption{Quantitative comparisons of the ablation study. \textcolor{definedPink}{Pink} and \textcolor{definedBlue}{blue} highlight the best and second-best results, respectively.}
\label{table:ab}
\end{table}

\begin{figure}[t!]
  \centering
  \includegraphics[width=0.95\linewidth]{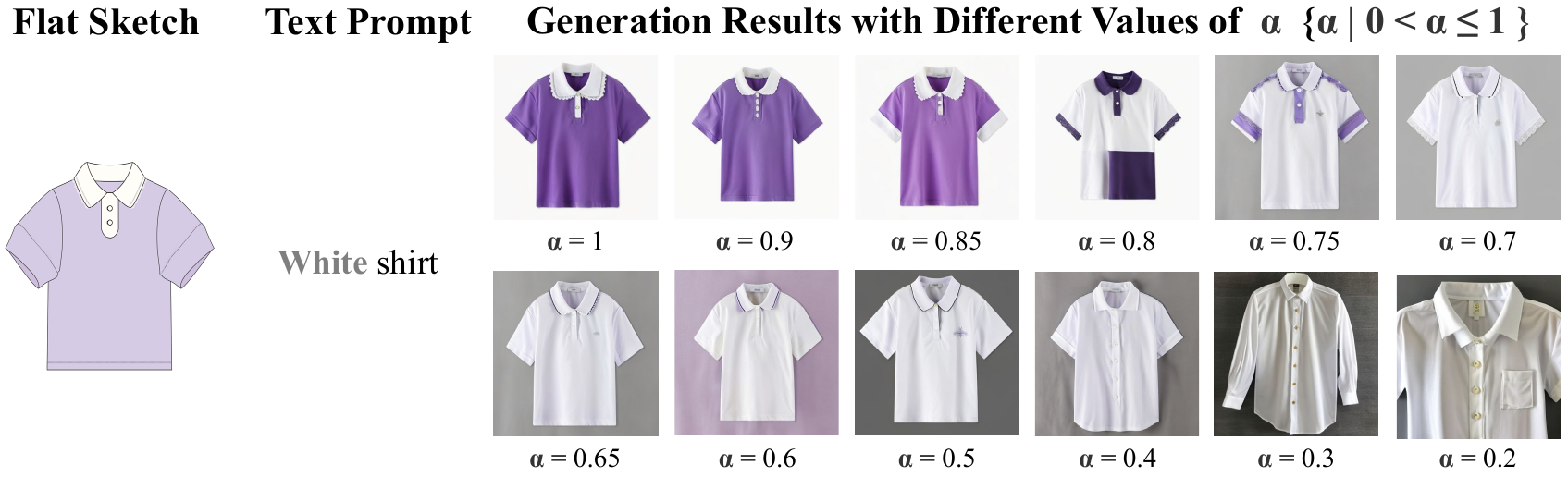}  
  \caption{Visual results of parameter analysis.} 
  \label{fig:lamda}
\end{figure}

\noindent\textbf{Parameter analysis.} We analyze the effect of parameter \(\alpha\) in Eq. \ref{eq:4}. We manually set the \(\alpha\) in the range between 0 to 1. As shown in Fig. \ref{fig:lamda}, with a higher \( \alpha_{\text{}} \) (e.g., \( \alpha_{\text{}} \) = 1), the HiGarment tends to generate an image as same as the reference image, and vice versa. When \( \alpha \) falls below 0.7, the generated image increasingly reflects the requirements specified in the text. However, when \( \alpha \) drops below 0.6, localized details from the flat sketch (patterns on the collar or button types) begin to diminish. Therefore, we set the parameter of the Sigmoid function \( \lambda \) as 0.6 to restrict \( \alpha \) ranges from 0.6 to 1 to keep a fidelity and accurate performance. 


\section{Conclusion and Limitations}
In this paper, we introduce Flat Sketch to Realistic Garment Image, a task for generating realistic garment images from flat sketches and text prompts. To address this task, we propose HiGarment, a multi-modal synthesis method that harmonizes text and image modalities. HiGarment incorporates a multi-modal semantic enhancement mechanism to strengthen fabric information in both visual and textual representations. Additionally, we introduce a harmonized cross-attention mechanism that fuses multi-modal features, enabling the generation model to dynamically balance whether to produce a text-biased or image-biased image. For evaluation and supplement, we release MMDGarment, a multi-modal fashion dataset. Experiments demonstrate HiGarment’s effectiveness, highlighting its potential to inspire advancements in image generation. 



\section*{Acknowledgements}
This research was supported by the National Natural Science Foundation of China (No.~62436009, 62276258), the Key Technology R\&D Program of Ningbo (No.~2023Z143, 2025Z028).

{
    \small
    \bibliographystyle{ieeenat_fullname}
    \bibliography{main}

\begin{thebibliography}{38}
\providecommand{\natexlab}[1]{#1}
\providecommand{\url}[1]{\texttt{#1}}
\expandafter\ifx\csname urlstyle\endcsname\relax
  \providecommand{\doi}[1]{doi: #1}\else
  \providecommand{\doi}{doi: \begingroup \urlstyle{rm}\Url}\fi

\bibitem[Aus et~al.(2021)Aus, Moora, Vihma, Unt, Kiisa, and Kapur]{aus2021designing}
Reet Aus, Harri Moora, Markus Vihma, Reimo Unt, Marko Kiisa, and Sneha Kapur.
\newblock Designing for circular fashion: integrating upcycling into conventional garment manufacturing processes.
\newblock \emph{Fashion and Textiles}, 8:\penalty0 1--18, 2021.

\bibitem[Chen et~al.(2024)Chen, Chen, Zhang, Liu, Wang, Zhou, Zhang, Zhou, Wan, and Sun]{chen2024mllm}
Dongping Chen, Ruoxi Chen, Shilin Zhang, Yinuo Liu, Yaochen Wang, Huichi Zhou, Qihui Zhang, Pan Zhou, Yao Wan, and Lichao Sun.
\newblock Mllm-as-a-judge: Assessing multimodal llm-as-a-judge with vision-language benchmark.
\newblock \emph{arXiv preprint arXiv:2402.04788}, 2024.

\bibitem[Chen et~al.(2023)Chen, Wen, Huang, Hu, and Peng]{chen2023foldgen}
Jia Chen, Yanfang Wen, Jin Huang, Xinrong Hu, and Tao Peng.
\newblock Foldgen: Multimodal transformer for garment sketch-to-photo generation.
\newblock In \emph{Computer Graphics International}, pages 455--466, 2023.

\bibitem[Choi et~al.(2021)Choi, Park, Lee, and Choo]{choi2021viton}
Seunghwan Choi, Sunghyun Park, Minsoo Lee, and Jaegul Choo.
\newblock Viton-hd: High-resolution virtual try-on via misalignment-aware normalization.
\newblock In \emph{CVPR}, pages 14131--14140, 2021.

\bibitem[Dhariwal and Nichol(2021)]{dhariwal2021diffusion}
Prafulla Dhariwal and Alexander Nichol.
\newblock Diffusion models beat gans on image synthesis.
\newblock \emph{NeurIPS}, 34:\penalty0 8780--8794, 2021.

\bibitem[Feng(2024)]{feng2024enhancing}
Chun-Mei Feng.
\newblock Enhancing label-efficient medical image segmentation with text-guided diffusion models.
\newblock \emph{arXiv preprint arXiv:2407.05323}, 2024.

\bibitem[Gupta et~al.(2023)Gupta, Kharbanda, Zhou, Li, Pfister, and Wei]{gupta2023cliptrans}
Devaansh Gupta, Siddhant Kharbanda, Jiawei Zhou, Wanhua Li, Hanspeter Pfister, and Donglai Wei.
\newblock Cliptrans: transferring visual knowledge with pre-trained models for multimodal machine translation.
\newblock In \emph{ICCV}, pages 2875--2886, 2023.

\bibitem[Han et~al.(2018)Han, Wu, Wu, Yu, and Davis]{han2018viton}
Xintong Han, Zuxuan Wu, Zhe Wu, Ruichi Yu, and Larry~S Davis.
\newblock Viton: An image-based virtual try-on network.
\newblock In \emph{CVPR}, pages 7543--7552, 2018.

\bibitem[He et~al.(2024)He, Yao, Zhang, Yu, Liu, and Xu]{he2024dresscode}
Kai He, Kaixin Yao, Qixuan Zhang, Jingyi Yu, Lingjie Liu, and Lan Xu.
\newblock Dresscode: Autoregressively sewing and generating garments from text guidance.
\newblock \emph{ACM Trans. Graph.}, 43\penalty0 (4):\penalty0 1--13, 2024.

\bibitem[Hessel et~al.(2021)Hessel, Holtzman, Forbes, Le~Bras, and Choi]{hessel2021clipscore}
Jack Hessel, Ari Holtzman, Maxwell Forbes, Ronan Le~Bras, and Yejin Choi.
\newblock Clipscore: A reference-free evaluation metric for image captioning.
\newblock In \emph{EMNLP}, pages 7514--7528, 2021.

\bibitem[Heusel et~al.(2017)Heusel, Ramsauer, Unterthiner, Nessler, and Hochreiter]{heusel2017gans}
Martin Heusel, Hubert Ramsauer, Thomas Unterthiner, Bernhard Nessler, and Sepp Hochreiter.
\newblock Gans trained by a two time-scale update rule converge to a local nash equilibrium.
\newblock \emph{NeurIPS}, 30:\penalty0 6629--6640, 2017.

\bibitem[Ho et~al.(2020)Ho, Jain, and Abbeel]{ho2020denoising}
Jonathan Ho, Ajay Jain, and Pieter Abbeel.
\newblock Denoising diffusion probabilistic models.
\newblock \emph{NeurIPS}, 33:\penalty0 6840--6851, 2020.

\bibitem[Kawano and Aoki(2024)]{kawano2024maskdiffusion}
Yasufumi Kawano and Yoshimitsu Aoki.
\newblock Maskdiffusion: Exploiting pre-trained diffusion models for semantic segmentation.
\newblock \emph{arXiv preprint arXiv:2403.11194}, 2024.

\bibitem[Li et~al.(2024)Li, Li, and Hoi]{li2024blip}
Dongxu Li, Junnan Li, and Steven Hoi.
\newblock Blip-diffusion: Pre-trained subject representation for controllable text-to-image generation and editing.
\newblock \emph{NeurIPS}, 36, 2024.

\bibitem[Li et~al.(2023)Li, Li, Savarese, and Hoi]{li2023blip}
Junnan Li, Dongxu Li, Silvio Savarese, and Steven Hoi.
\newblock Blip-2: Bootstrapping language-image pre-training with frozen image encoders and large language models.
\newblock In \emph{ICML}, pages 19730--19742, 2023.

\bibitem[Lin et~al.(2022)Lin, Chen, Xu, Zhang, Lu, and Zhang]{lin2022ds}
Ailiang Lin, Bingzhi Chen, Jiayu Xu, Zheng Zhang, Guangming Lu, and David Zhang.
\newblock Ds-transunet: Dual swin transformer u-net for medical image segmentation.
\newblock \emph{IEEE Transactions on Instrumentation and Measurement}, 71:\penalty0 1--15, 2022.

\bibitem[Liu et~al.(2023)Liu, Wang, Liu, He, and Liu]{liu2023ed}
Jiawei Liu, Weining Wang, Wei Liu, Qian He, and Jing Liu.
\newblock Ed-t2v: An efficient training framework for diffusion-based text-to-video generation.
\newblock In \emph{IJCNN}, pages 1--8, 2023.

\bibitem[Liu et~al.(2024)Liu, Wang, Fan, Wang, Tang, and Qu]{liu2024residual}
Jiawei Liu, Qiang Wang, Huijie Fan, Yinong Wang, Yandong Tang, and Liangqiong Qu.
\newblock Residual denoising diffusion models.
\newblock In \emph{CVPR}, pages 2773--2783, 2024.

\bibitem[Loshchilov(2017)]{loshchilov2017decoupled}
I Loshchilov.
\newblock Decoupled weight decay regularization.
\newblock \emph{arXiv preprint arXiv:1711.05101}, 2017.

\bibitem[Ma et~al.(2024)Ma, Li, and Wang]{ma2024u}
Jun Ma, Feifei Li, and Bo Wang.
\newblock U-mamba: Enhancing long-range dependency for biomedical image segmentation.
\newblock \emph{arXiv preprint arXiv:2401.04722}, 2024.

\bibitem[Morelli et~al.(2022)Morelli, Fincato, Cornia, Landi, Cesari, and Cucchiara]{morelli2022dress}
Davide Morelli, Matteo Fincato, Marcella Cornia, Federico Landi, Fabio Cesari, and Rita Cucchiara.
\newblock Dress code: High-resolution multi-category virtual try-on.
\newblock In \emph{CVPR}, pages 2231--2235, 2022.

\bibitem[Mou et~al.(2024)Mou, Wang, Xie, Wu, Zhang, Qi, and Shan]{mou2024t2i}
Chong Mou, Xintao Wang, Liangbin Xie, Yanze Wu, Jian Zhang, Zhongang Qi, and Ying Shan.
\newblock T2i-adapter: Learning adapters to dig out more controllable ability for text-to-image diffusion models.
\newblock In \emph{AAAI}, pages 4296--4304, 2024.

\bibitem[Qi et~al.(2024)Qi, Fang, Wu, Xie, Liu, Chen, He, and Zhang]{qi2024deadiff}
Tianhao Qi, Shancheng Fang, Yanze Wu, Hongtao Xie, Jiawei Liu, Lang Chen, Qian He, and Yongdong Zhang.
\newblock Deadiff: An efficient stylization diffusion model with disentangled representations.
\newblock In \emph{CVPR}, pages 8693--8702, 2024.

\bibitem[Radford et~al.(2021)Radford, Kim, Hallacy, Ramesh, Goh, Agarwal, Sastry, Askell, Mishkin, Clark, et~al.]{radford2021learning}
Alec Radford, Jong~Wook Kim, Chris Hallacy, Aditya Ramesh, Gabriel Goh, Sandhini Agarwal, Girish Sastry, Amanda Askell, Pamela Mishkin, Jack Clark, et~al.
\newblock Learning transferable visual models from natural language supervision.
\newblock In \emph{ICML}, pages 8748--8763, 2021.

\bibitem[Rombach et~al.(2022)Rombach, Blattmann, Lorenz, Esser, and Ommer]{rombach2022high}
Robin Rombach, Andreas Blattmann, Dominik Lorenz, Patrick Esser, and Bj{\"o}rn Ommer.
\newblock High-resolution image synthesis with latent diffusion models.
\newblock In \emph{CVPR}, pages 10684--10695, 2022.

\bibitem[Sauer et~al.(2025)Sauer, Lorenz, Blattmann, and Rombach]{sauer2025adversarial}
Axel Sauer, Dominik Lorenz, Andreas Blattmann, and Robin Rombach.
\newblock Adversarial diffusion distillation.
\newblock In \emph{ECCV}, pages 87--103, 2025.

\bibitem[Sohl-Dickstein et~al.(2015)Sohl-Dickstein, Weiss, Maheswaranathan, and Ganguli]{sohl2015deep}
Jascha Sohl-Dickstein, Eric Weiss, Niru Maheswaranathan, and Surya Ganguli.
\newblock Deep unsupervised learning using nonequilibrium thermodynamics.
\newblock In \emph{ICML}, pages 2256--2265, 2015.

\bibitem[Wan et~al.(2014)Wan, Mok, and Jin]{wan2014shape}
Xianmei Wan, Pik~Yin Mok, and Xiaogang Jin.
\newblock Shape deformation using skeleton correspondences for realistic posed fashion flat creation.
\newblock \emph{IEEE Transactions on Automation Science and Engineering}, 11\penalty0 (2):\penalty0 409--420, 2014.

\bibitem[Xu et~al.(2023)Xu, Wang, Zhang, Wang, and Shi]{xu2023versatile}
Xingqian Xu, Zhangyang Wang, Gong Zhang, Kai Wang, and Humphrey Shi.
\newblock Versatile diffusion: Text, images and variations all in one diffusion model.
\newblock In \emph{ICCV}, pages 7754--7765, 2023.

\bibitem[Ye et~al.(2023)Ye, Zhang, Liu, Han, and Yang]{ye2023ip}
Hu Ye, Jun Zhang, Sibo Liu, Xiao Han, and Wei Yang.
\newblock Ip-adapter: Text compatible image prompt adapter for text-to-image diffusion models.
\newblock \emph{arXiv preprint arXiv:2308.06721}, 2023.

\bibitem[Zhang et~al.(2023{\natexlab{a}})Zhang, Rao, and Agrawala]{zhang2023adding}
Lvmin Zhang, Anyi Rao, and Maneesh Agrawala.
\newblock Adding conditional control to text-to-image diffusion models.
\newblock In \emph{ICCV}, pages 3836--3847, 2023{\natexlab{a}}.

\bibitem[Zhang et~al.(2018)Zhang, Isola, Efros, Shechtman, and Wang]{zhang2018unreasonable}
Richard Zhang, Phillip Isola, Alexei~A Efros, Eli Shechtman, and Oliver Wang.
\newblock The unreasonable effectiveness of deep features as a perceptual metric.
\newblock In \emph{CVPR}, pages 586--595, 2018.

\bibitem[Zhang et~al.(2024{\natexlab{a}})Zhang, Chong, Zhang, Li, Cheng, Yan, and Liang]{zhang2024garmentaligner}
Shiyue Zhang, Zheng Chong, Xujie Zhang, Hanhui Li, Yuhao Cheng, Yiqiang Yan, and Xiaodan Liang.
\newblock Garmentaligner: Text-to-garment generation via retrieval-augmented multi-level corrections.
\newblock \emph{arXiv preprint arXiv:2408.12352}, 2024{\natexlab{a}}.

\bibitem[Zhang et~al.(2022)Zhang, Sha, Kampffmeyer, Xie, Jie, Huang, Peng, and Liang]{zhang2022armani}
Xujie Zhang, Yu Sha, Michael~C Kampffmeyer, Zhenyu Xie, Zequn Jie, Chengwen Huang, Jianqing Peng, and Xiaodan Liang.
\newblock Armani: Part-level garment-text alignment for unified cross-modal fashion design.
\newblock In \emph{ACM MM}, pages 4525--4535, 2022.

\bibitem[Zhang et~al.(2023{\natexlab{b}})Zhang, Yang, Kampffmeyer, Zhang, Zhang, Lu, Lin, Xu, and Liang]{zhang2023diffcloth}
Xujie Zhang, Binbin Yang, Michael~C Kampffmeyer, Wenqing Zhang, Shiyue Zhang, Guansong Lu, Liang Lin, Hang Xu, and Xiaodan Liang.
\newblock Diffcloth: Diffusion based garment synthesis and manipulation via structural cross-modal semantic alignment.
\newblock In \emph{ICCV}, pages 23154--23163, 2023{\natexlab{b}}.

\bibitem[Zhang et~al.(2024{\natexlab{b}})Zhang, Song, Liu, Wang, Yu, Tang, Li, Tang, Hu, Pan, et~al.]{zhang2024ssr}
Yuxuan Zhang, Yiren Song, Jiaming Liu, Rui Wang, Jinpeng Yu, Hao Tang, Huaxia Li, Xu Tang, Yao Hu, Han Pan, et~al.
\newblock Ssr-encoder: Encoding selective subject representation for subject-driven generation.
\newblock In \emph{CVPR}, pages 8069--8078, 2024{\natexlab{b}}.

\bibitem[Zhang et~al.(2024{\natexlab{c}})Zhang, Zhang, and Xie]{zhang2024texcontrol}
Yongming Zhang, Tianyu Zhang, and Haoran Xie.
\newblock Texcontrol: Sketch-based two-stage fashion image generation using diffusion model.
\newblock \emph{arXiv preprint arXiv:2405.04675}, 2024{\natexlab{c}}.

\bibitem[Zhao et~al.(2024)Zhao, Chen, Chen, Bao, Hao, Yuan, and Wong]{zhao2024uni}
Shihao Zhao, Dongdong Chen, Yen-Chun Chen, Jianmin Bao, Shaozhe Hao, Lu Yuan, and Kwan-Yee~K Wong.
\newblock Uni-controlnet: All-in-one control to text-to-image diffusion models.
\newblock \emph{NeurIPS}, 36, 2024.

\end{thebibliography}
}
\clearpage
\setcounter{page}{1}
\maketitlesupplementary

\section{Dataset details}
\label{sec:dataset}
\subsection{Dataset construction}
\begin{figure*}[htbp]
\centering
\includegraphics[width=1\textwidth]{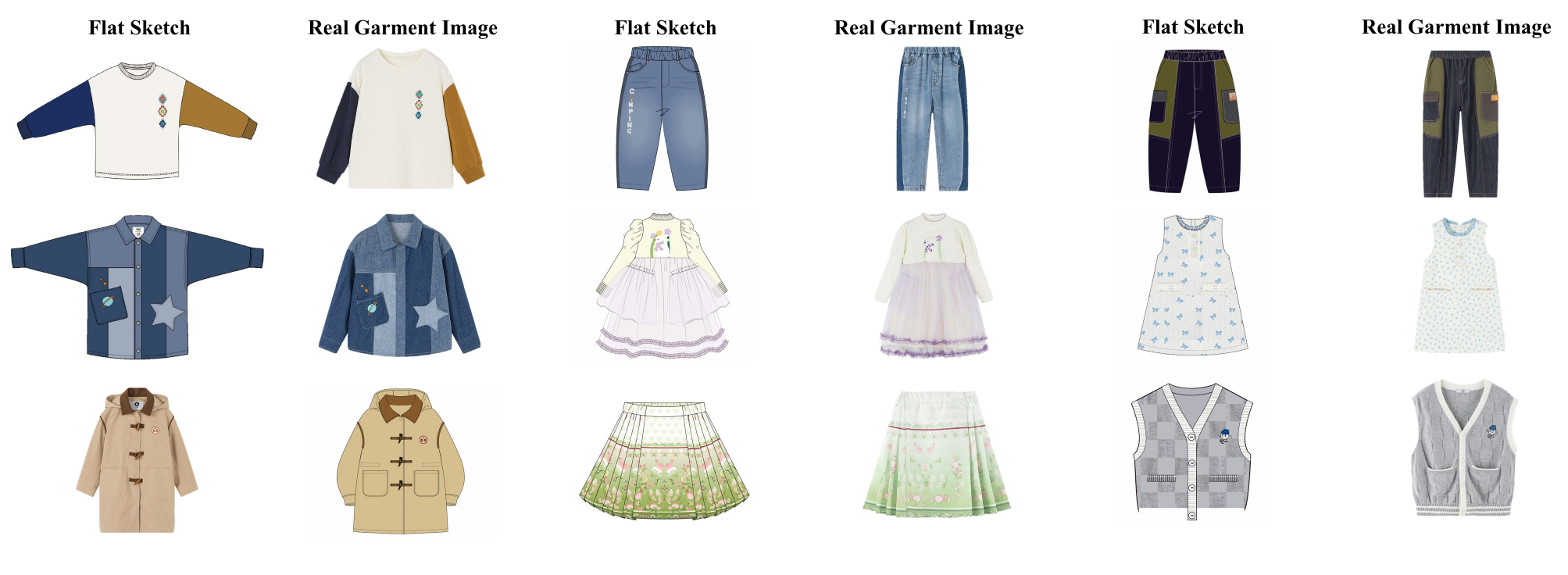}
\caption{Samples of the flat sketches and corresponding real garment images in the collected MMDGarment dataset.}\label{fig:dataset1}
\end{figure*}

\begin{figure*}[htbp]
\centering
\includegraphics[width=1\textwidth]{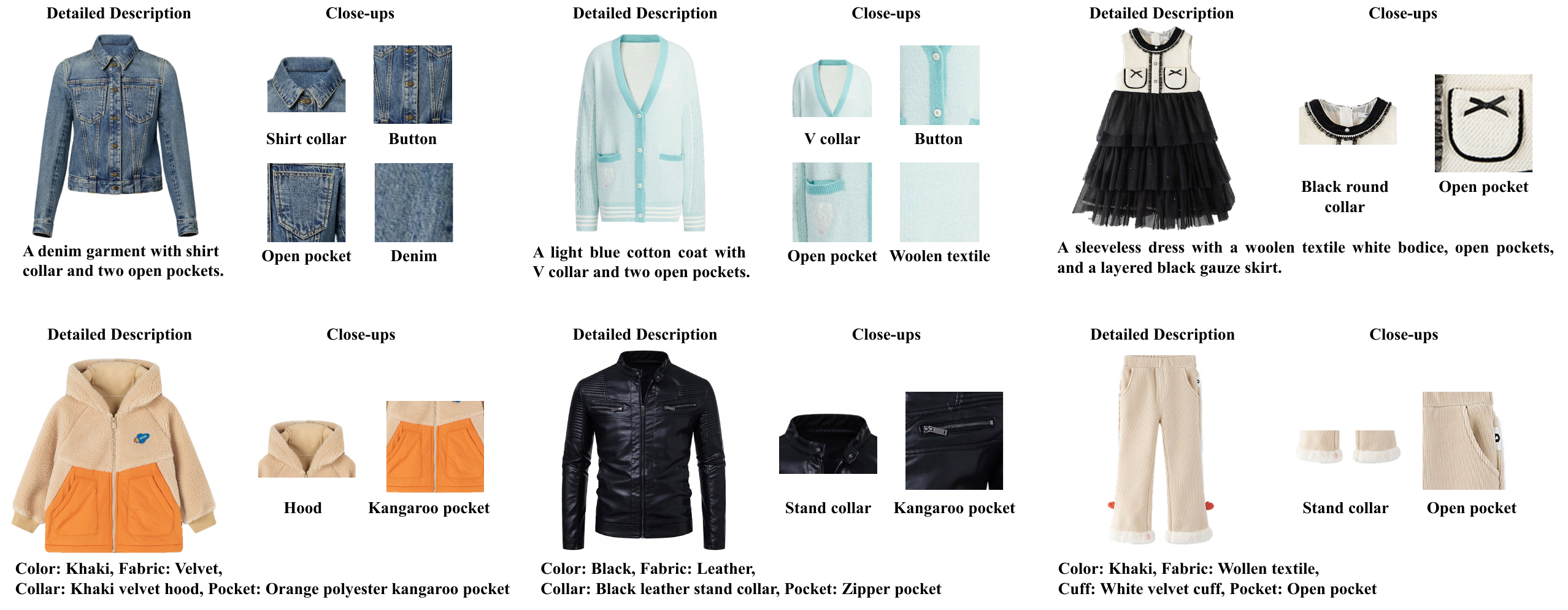}
\caption{Samples of the detailed description and close-ups in the collected MMDGarment dataset.}\label{fig:dataset2}
\end{figure*}

The dataset collection team comprised six members, each assigned to three to five garment brand websites. Images were collected from multiple garment brand websites, with specific requirements to ensure clear display of fabric and color characteristics.  Before official collection, all team members were required to carefully study a standardized tutorial and strictly follow the tutorial’s guidelines during the collection process. Images were collected from multiple garment brand websites, with specific requirements to ensure clear display of fabric and color characteristics. For each site, members adhered to predefined standards for image resolution, file format, and garment presentation to ensure consistent image quality across the dataset. Only images with a sufficiently clear display of fabric and color characteristics were selected to enable accurate identification of garment details. The annotation strategy required single-color garments to be labeled with precise color information, whereas garments with complex colors focused on fabric and structural annotations. To guarantee the reliability of the annotations, the team regularly consulted professional garment designers, who provided expert verification of fabric types and other visual features. Fig. \ref{fig:dataset1} and Fig. \ref{fig:dataset2} provide some samples of flat sketches, close-ups, and detailed descriptions from the MMDGarment dataset.

For the fabric vector database, the team also downloaded high-resolution sample images exhibiting distinct fabric characteristics, curated according to fabric types frequently encountered in professional fashion design practice. To ensure practical relevance, the covered fabric types were chosen based on frequency and importance in professional fashion design practice, as identified through consultation with experienced designers and review of contemporary fashion collections. The resulting fabric vector database consists of 150 image-text pairs, covering 11 major fabric categories (e.g., cotton, wool, denim, silk, lace, and synthetic blends), thereby providing a comprehensive and diverse reference for fabric representation tasks.
\subsection{Dataset analysis}
In addition to the main dataset, the MMDGarment also includes 1,975 close-up images of garment details, such as collars, sleeves, and pockets, with detailed fabric and color annotations essential for training models to capture intricate garment features. The flat sketches are technical drawings of real garment designs, created by professional designers from our collaborating garment company. Tab. \ref{tab:dataset_comp} demonstrates a comparison between MMDGarment and other mainstream fashion datasets \cite{zhang2022armani, morelli2022dress, choi2021viton}

\begin{table}[h!]
\centering
\renewcommand{\arraystretch}{1.8} 
\scalebox{0.46}{ 
\begin{tabular}{@{}m{2.5cm} c c c c c c c m{3.35cm}@{}}
\toprule
\textbf{Datasets} & \textbf{Public} & \textbf{Close-ups} & \textbf{Fabric} & \textbf{Color} & \textbf{Sketch} & \textbf{\# Pairs} & \textbf{Example Image} & \textbf{Example Text} \\ 
\midrule
{ CM-Fashion \cite{zhang2022armani}} & {\Large \ding{55}} & {\Large \checkmark} & {\Large \checkmark} & {\Large \checkmark} & {\Large \ding{55}} & {\Large 500,000} & \raisebox{-0.5\height}{\includegraphics[width=2cm]{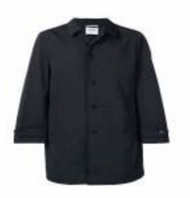}} & { classic collar Navy blue cotton blend button-up trench coat} \\
{ DressCode \cite{morelli2022dress}} & {\Large \checkmark} & {\Large \ding{55}} & {\Large \ding{55}} & {\Large \ding{55}} & {\Large \ding{55}} & {\Large 53,795} & \raisebox{-0.5\height}{\includegraphics[width=2cm]{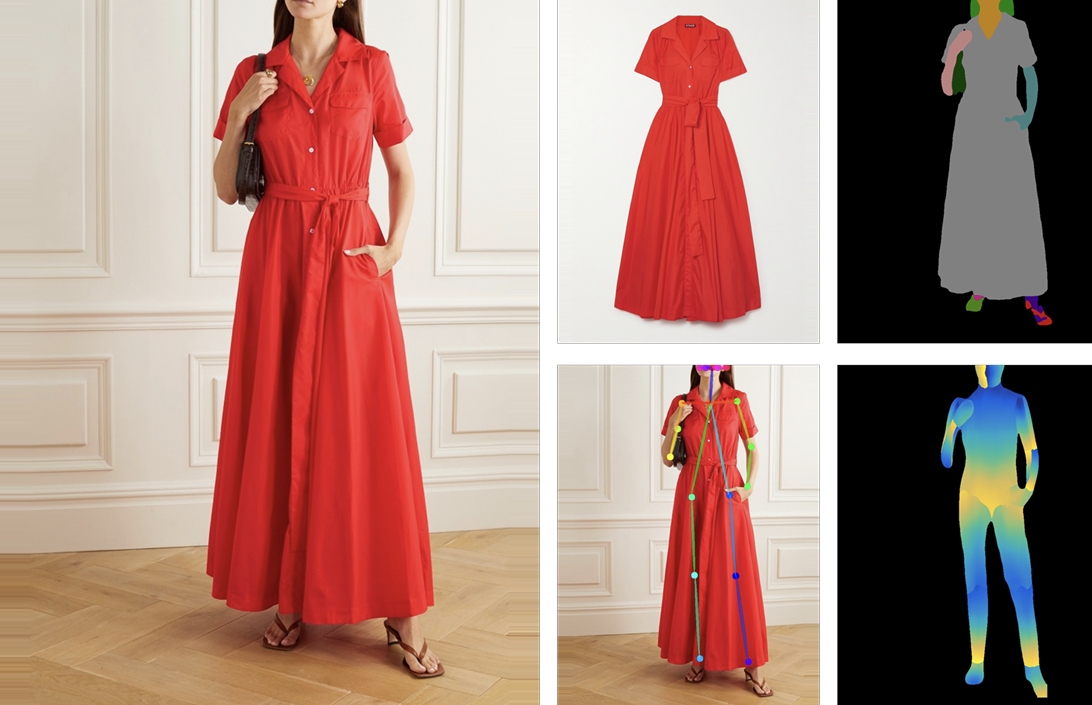}} & { (Body key points and segmentation masks are provided)} \\
{ VITON-HD \cite{choi2021viton}} & {\Large \checkmark} & {\Large \ding{55}} & {\Large \checkmark} & {\Large \checkmark} & {\Large \ding{55}} & {\Large 13,679} & \raisebox{-0.5\height}{\includegraphics[width=2cm]{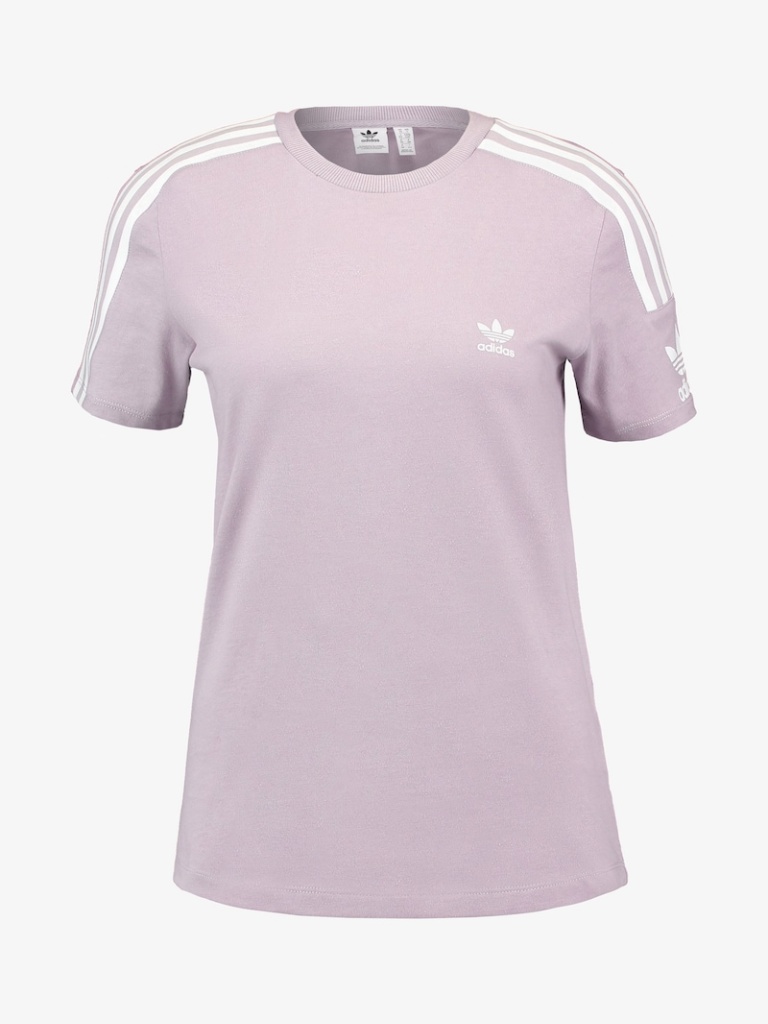}} & { Embroidery Lavender lettering prints Cotton Short Sleeve normal-fit Round Neck T-shirts} \\
{ \textbf{MMDGarment (Ours)}} & {\Large \checkmark} & {\Large \checkmark} & {\Large \checkmark} & {\Large \checkmark} & {\Large \checkmark} & {\Large 20,151} & \raisebox{-0.5\height}{\includegraphics[width=2cm]{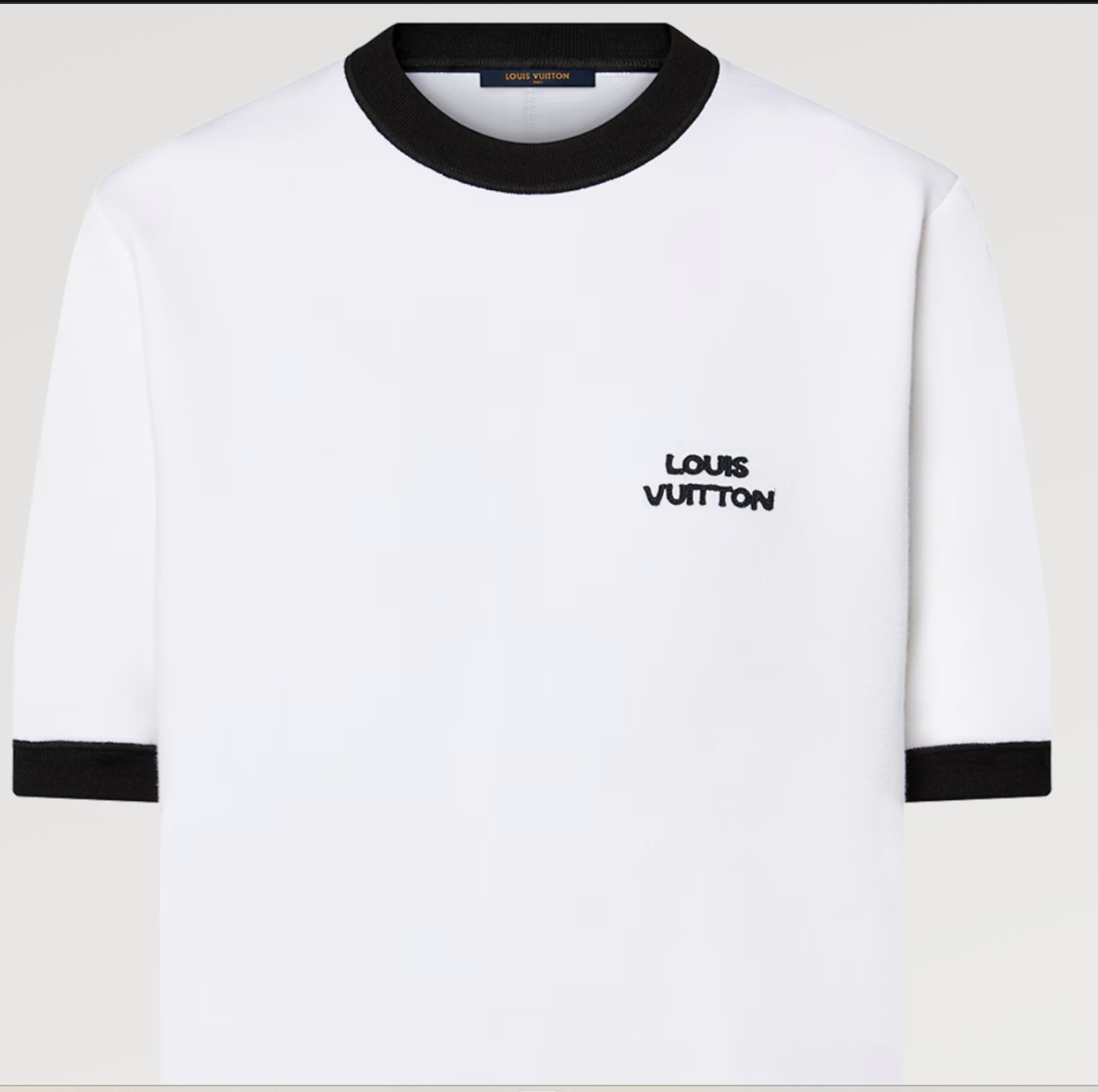}} & { White cotton t-shirt with a black round collar, white short sleeves, black sleeve cuff.} \\
\bottomrule
\end{tabular}
}
\caption{Comparison of the most widely used datasets for garment synthesis tasks}
\label{tab:dataset_comp}
\end{table}


\section{Visualization}
\label{sec:vis}
\begin{figure}[h!]
  \centering
  \includegraphics[width=0.95\linewidth]{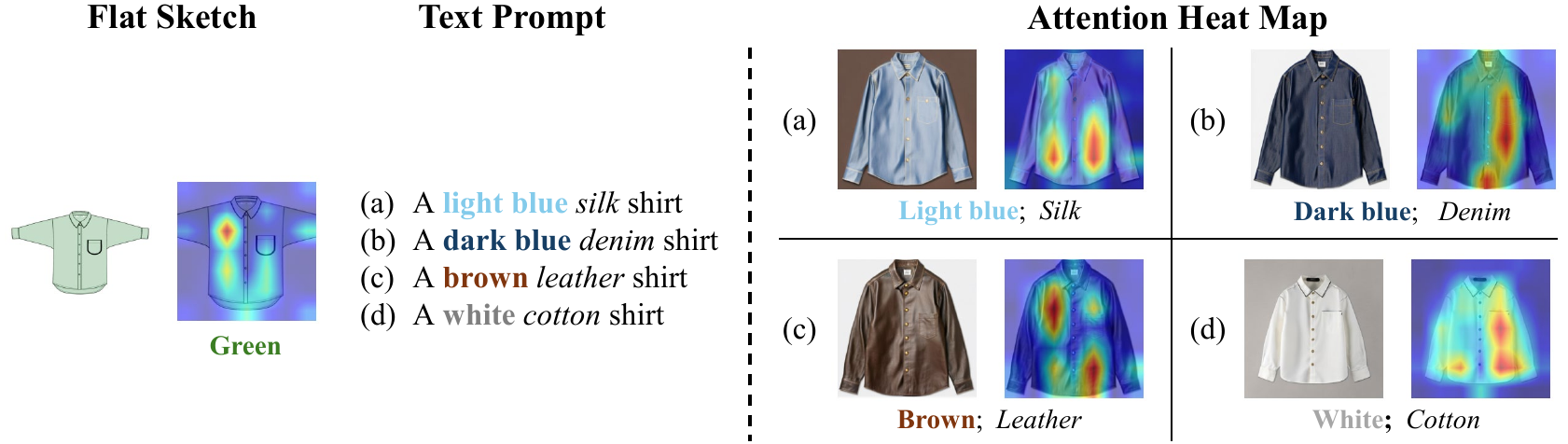}  
  \caption{Cross-attention visualization for generated garment images with different color-fabric attribute composition.} 
  \label{fig:vis1}
\end{figure}

\begin{figure}[h]
  \centering
  \includegraphics[width=0.9\linewidth]{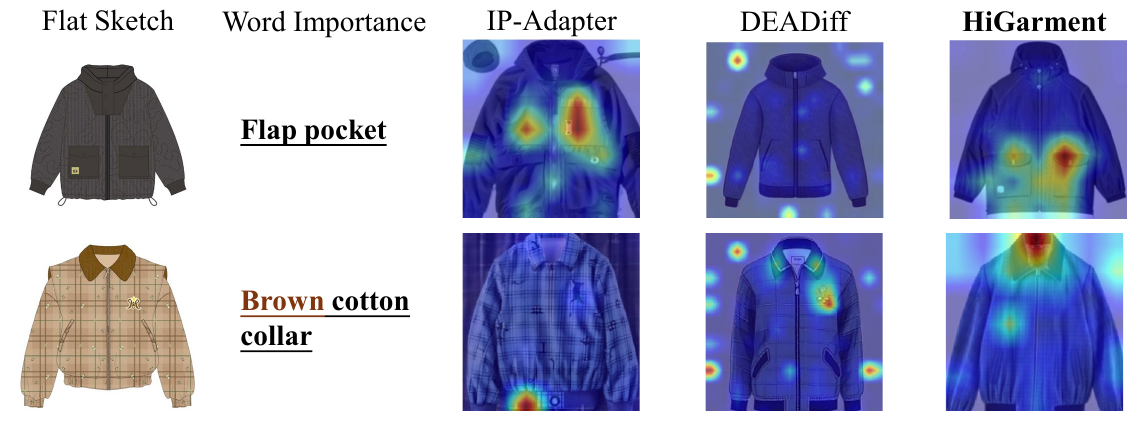}  
  \caption{Cross-attention visualization comparison between HiGarment and other methods.} 
  \label{fig:attword}
\end{figure}

We conducted a series of visualization experiments on attention heat maps to demonstrate the effectiveness of our method in accurately generating garment components and attributes. Fig. \ref{fig:vis1} shows that the generated garment images capture attribute details, such as color and fabric, as specified in the textual input. We also compare the cross-attention visualization results between other methods and HiGarment as shown in Fig. \ref{fig:attword}. Our method demonstrates significantly more distinct attention to key components such as the pocket and collar. These results validate the critical role of modality harmony in addressing the FS2RG task.

\section{MLLM-based evaluation and user study}
\label{sec:mllm}

The MLLM evaluation contains four aspects: structure (35\%), color (25\%), fabric (25\%), and details (15\%). Specifically, structure refers to the overall layout and key components of the garment, such as the shape, style, and arrangement of major parts (e.g., collars, sleeves, and hoods), demonstrating the model’s ability to reconstruct the correct garment shape and structural attributes. Color evaluates the accuracy and consistency of the generated garment’s colors with respect to the reference, reflecting the model’s capability in color reproduction and harmony. Fabric assesses whether the generated images correctly represent the type, texture, and appearance of the garment materials, thus verifying the model’s ability to capture subtle material differences. Details focus on fine-grained elements, including stitching, decorative features, and small structural or visual cues, highlighting the model’s proficiency in generating intricate garment details.

Each aspect is scored from 0 to 10 by the MLLM, with higher scores indicating better fidelity to the ground truth in that dimension. The lowest possible score for a pair is 0 (if all aspects score 0), and the highest possible score is 10 (if all aspects receive the maximum score). These four scores are then combined using a weighted sum to produce a final score for each pair, which allows for a comprehensive assessment of the generated images across both global and local garment characteristics. For the evaluation process, we first selected 28 pairs of generated images and corresponding product images. For each pair, the associated instruction was provided to the MLLM as a prompt, together with both the generated and real product images. The MLLM was tasked to rate each pair independently on structure, color, fabric, and details according to predefined evaluation criteria. After scoring all four aspects for each pair, the weighted sum was calculated to obtain the pair’s overall score. Finally, the final evaluation result is reported as the average score across all 28 pairs, providing an objective and multi-faceted measure of the model’s performance in garment generation. 

For the user study, we invited 2 professional garment designers from the corporate company, 11 fashion design students from NingboTech University, and 20 non-experts majoring in computer science to assess the similarity between generated garment images and real garment photos. Each participant was provided with a questionnaire that mirrored the evaluation criteria used in the MLLM assessment, covering four aspects: structure, color, fabric, and details. For each image pair, the participants were asked to score each aspect on a scale from 0 to 10, following clear evaluation guidelines. This human evaluation not only offers a subjective perspective complementary to the automated MLLM assessment but also enables us to examine the consistency between expert, trained, and layperson judgments. The strong alignment between MLLM scores and human ratings demonstrates the effectiveness of using large models as evaluators in this task. Moreover, the high scores achieved by our method in both objective (MLLM-based) and subjective (human-based) evaluations validate its robustness and reliability from both technical and perceptual standpoints.

\begin{figure}[h]
  \centering
  \includegraphics[width=0.9\linewidth]{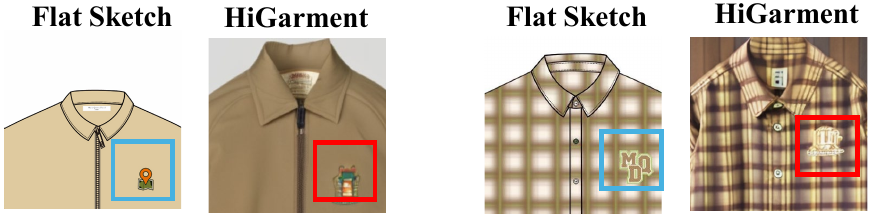}  
  \caption{Failure cases for tiny logoes.} 
  \label{fig:fail}
\end{figure}

\section{Limitations and failure cases}
\label{sec:limitation}
We analyze the limitations and failure cases in this section. Current evaluation remains limited to design-stage synthesis due to scarce public datasets and garment generation codes. We will expand comparisons when resources \cite{zhang2023diffcloth, zhang2024garmentaligner, chen2023foldgen} permit. Fig. \ref{fig:fail} illustrates several representative failure cases observed in HiGarment, particularly in preserving tiny logos and intricate garment details. These fine elements are often lost during the diffusion denoising process, leading to visible discrepancies between the generated images and the ground truth products. To address these issues in the future, we plan to explore the integration of specialized modules or attention mechanisms designed to enhance the preservation of fine-grained patterns in the diffusion process.

\end{document}